\documentclass{article}

\usepackage{microtype}
\usepackage{multirow}
\usepackage{graphicx}
\usepackage{subfigure}
\usepackage{booktabs} 

\usepackage{hyperref}

\usepackage[accepted]{icml_fmwild}

\usepackage{amsmath}
\usepackage{amssymb}
\usepackage{mathtools}
\usepackage{amsthm}
\usepackage{graphicx}
\usepackage{subcaption}

\usepackage[capitalize,noabbrev]{cleveref}

\theoremstyle{plain}

\theoremstyle{definition}

\theoremstyle{remark}

\usepackage[disable,textsize=tiny]{todonotes}
\usepackage[textsize=tiny]{todonotes}

\begin{document}

\twocolumn[
\icmltitle{Inference Performance Optimization for
	Large Language Models on CPUs}




\begin{icmlauthorlist}
\icmlauthor{Pujiang He}{comp}
\icmlauthor{Shan Zhou}{comp}
\icmlauthor{Wenhuan Huang}{comp}
\icmlauthor{Changqing Li}{comp}
\icmlauthor{Duyi Wang}{comp}
\icmlauthor{Bin Guo}{comp}
\icmlauthor{Chen Meng}{comp}
\icmlauthor{Sheng Gui}{comp}
\icmlauthor{Weifei Yu}{comp}
\icmlauthor{Yi Xie}{comp}

\end{icmlauthorlist}

\icmlaffiliation{comp}{Intel Corporation, Shanghai, China}
\icmlcorrespondingauthor{Pujiang He}{pujiang.he@intel.com}
\icmlcorrespondingauthor{Shan Zhou}{shan.zhou@intel.com}
\icmlcorrespondingauthor{Wenhuan Huang}{wenhuan.huang@intel.com}

\icmlkeywords{Large Language Models; Distributed Inferencen; Attention Optimization; x86 Processor}

\vskip 0.3in
]



\printAffiliationsAndNotice{}  

\begin{abstract}
		
Large language models (LLMs) have shown exceptional performance and vast potential across diverse tasks. However, the deployment of LLMs with high performance in low-resource environments has garnered significant attention in the industry. When GPU hardware resources are limited, we can explore alternative options on CPUs. To mitigate the financial burden and alleviate constraints imposed by hardware resources, optimizing inference performance is necessary. In this paper, we introduce an easily deployable inference performance optimization solution aimed at accelerating LLMs on CPUs. In this solution, we implement an effective way to reduce the KV cache size while ensuring precision. We propose a distributed inference optimization approach and implement it based on oneAPI Collective Communications Library. Furthermore, we propose optimization approaches for LLMs on CPU, and conduct tailored optimizations for the most commonly used models. 
The code is open-sourced at  \url{https://github.com/intel/xFasterTransformer}.
		
\end{abstract}
	
\section{Introduction}
Large language models (LLMs) based on the Transformer \citet{vaswani2017attention} architecture have garnered profound technical attention globally and achieved remarkable accomplishments 
	\cite{touvron2023llama} , 
	\cite{zhang2022opt}, 
	\cite{yang2023baichuan}, 
	\cite{wu2023brief},
	\cite{bai2023qwen}.
Their robust comprehension and generation capabilities are profoundly changing applications of artificial intelligence \cite{thirunavukarasu2023large}. 
However, the practical deployment of LLMs is significantly hindered by the high cost and resource limitations of hardware \cite{zhao2023survey}. 
Therefore, deploying of LLM with good performance for practical applications has become a trending topic in industry, which helps land LLMs in more practical applications. When GPU hardware resources are limited, we can explore alternative options on CPUs.
	
Optimizing the practical deployment performance of LLMs necessitates effectively leveraging hardware capabilities for cost efficiency and improved inference performance, which, in turn, demands that developers possess a strong understanding of both hardware and software.  
To tackle the practical deployment challenge, we propose an easy and efficient solution designed to facilitate easy deployment on CPUs. 
Deploying on CPUs offers the advantage of being unrestricted by VRAM size, preventing KV cache overflow, and enabling the processing of extremely long-context support \cite{liu2023lost}.
Furthermore, deployment on CPUs can enhance system resource utilization, making multitasking more efficient.

The paper introduces the solution for efficient deployment and inference performance optimization for LLMs on CPUs. The solution supports for widely used LLMs, and
experiment results show the proposed solution has good inference performance scalability on CPUs. 
We have established a repository where we curate relevant optimization with real-time updates.
The main contributions are as follows:

\begin{itemize}

	\item We propose new LLM optimize solutions on CPUs, such as SlimAttention. We conduct individual optimizations for LLM operations and layers, and the support extends to widely used LLMs, encompassing Qwen, Llama, ChatGLM , Baichuan, and Opt series.
	
	\item We implement an effective way to reduce the KV cache size and ensure precision. This approach allows for a more efficient use of memory without significantly compromising the quality of the model's output.
	
	\item We design a distributed inference optimization solution for LLMs on CPUs, facilitating the attainment of necessary scalability and efficient low-latency inference.

\end{itemize}

\section{Approach}
\label{approach}
	
In this section, three main optimization approaches are proposed and more details could be referred to the following sections.

\subsection{LLM Optimization}
To improve the inference performance, we proposed individual optimize solutions for each LLM operations and layers. Taking attention layer as an example. 
Since the resources consumed by the attention mechanism are directly proportional to the square of the sequence length from theoretical,  optimizing attention is particularly important for long sequence inputs.

\begin{figure}
			\centering
			\includegraphics[width=.8\linewidth]{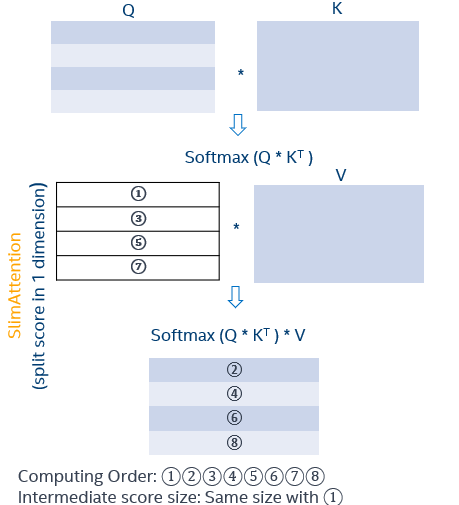}  
			\caption{Slim Attention}
			\label{attention opt}
\end{figure}

\begin{figure}
			\centering
			\includegraphics[width=.8\linewidth]{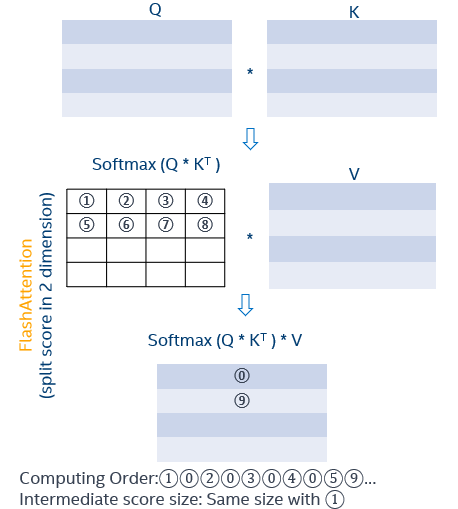}  
			\caption{Flash Attention}
			\label{Flash Attention}
\end{figure}

A new approach we called SlimAttention shown in Figure ~\ref{attention opt} is proposed. SlimAttention is essentially a one-dimensional decomposition of the score between query and key.
In terms of computation sequence, it involves first calculating a horizontal score, followed by applying softmax to the score. 
The resulting softmax is then multiplied by the corresponding values, producing a portion of the output. 
This process is repeated for the next block using the same score buffer. 
In theory, each thread needs only to maintain a buffer of the block size, thereby reducing memory usage and enhancing computational efficiency in practice. 
And FlashAttention \cite{dao2022flashattention} is a solution initially introduced on GPUs, shown in Figure ~\ref{Flash Attention}. Essentially, it involves a two-dimensional decomposition of the score. In the computational process, each thread only needs to maintain a single tile. However, as a tile doesn't encompass all the data in the softmax direction, there is a need for iterative corrections to the final results during the calculation.  
	
Compared with FlashAttention, SlimAttention entails no redundant computations but does necessitate a larger intermediate buffer.

\subsection{Effective KV cache Optimization}
Typical LLM architectures are predominantly decoder-only structures. Within these models, computation is primarily divided into two categories: attention (MHA/GQA) and matrix multiplication (MatMul), the latter often including post-operations that can be fused into the MatMul process. During the generation of the first token, both attention and MatMul operations are compute-bound. However, for the generation of subsequent tokens (beyond the first token), the attention typically exhibits a pattern of general matrix-vector multiplication (gemv), which is bound in memory bandwidth.

The volume of the key-value (KV) cache accessed during the generation of a single token can be calculated as follows: 
\begin{equation}
	2  b  (L_i + L_o) l n_{head}  s_{head}    s_d
\end{equation}

where the b is batch size, $L_i$ and $L_o$ are input sequence length and output sequence length, $n_{head}$, $s_{head}$ $l$, an d$s_d$ respectively represents head number, head size,  layer, and size of the data type, and factor of 2 accounts for both the key and value components.

Consider the Llama2-7B model as an illustrative example, which is characterized by the following parameters: head\_number=32, head\_size=128, and layers=32. Assuming a batch size of 256, an input sequence length of 1024, an output sequence length of 1024, and the use of FP16/BF16 data types for both weights and the KV cache, the maximum KV cache required is approximately 128GB. In contrast, the weight data that needs to be accessed is only around 14GB. This comparison underscores the significance of the KV cache size during inference with large batch sizes.

\begin{figure}[h]
	\begin{center}
		\includegraphics[width=0.5\textwidth]{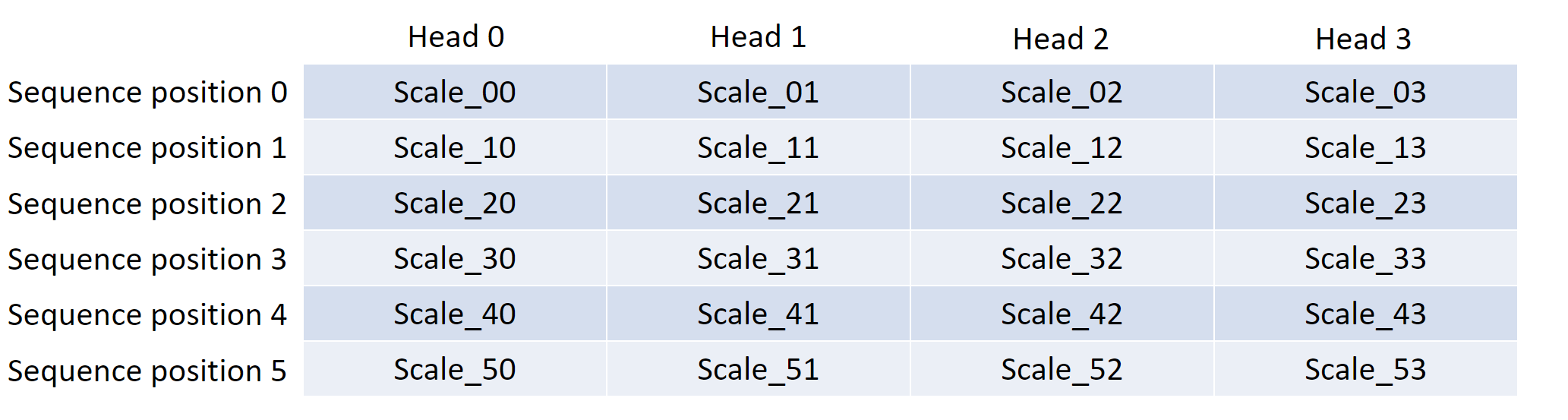}
	\end{center}
	\caption{Example of maintained unique scale.}
	\label{kvcache1}
\end{figure}

\begin{figure}[h]
	\begin{center}
		\includegraphics[width=0.5\textwidth]{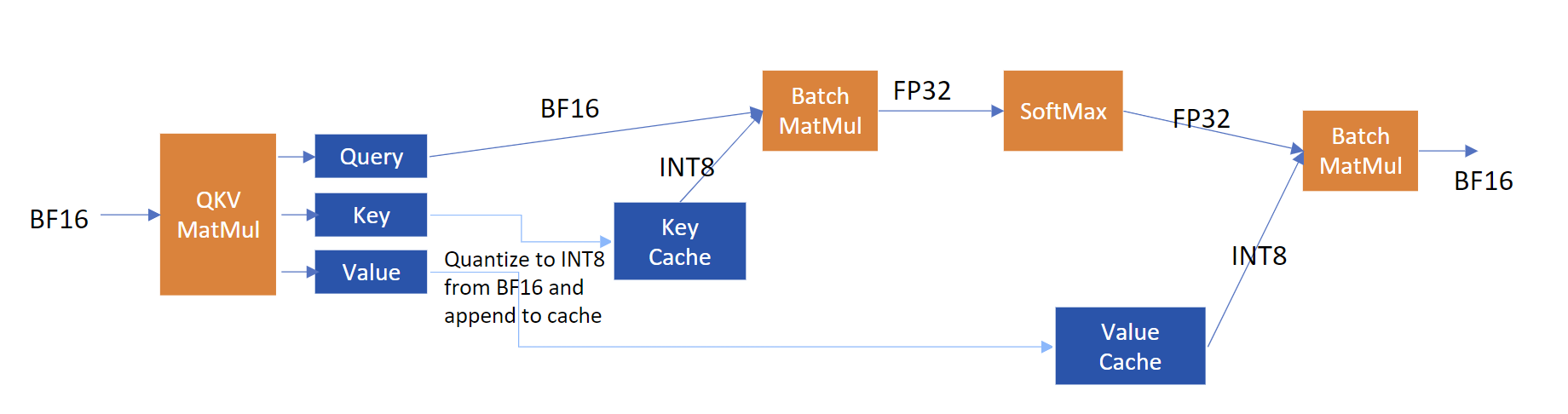}
	\end{center}
	\caption{Custom kernel workflow.}
	\label{kvcache2}
\end{figure}

To enhance performance by effectively reducing the KV cache size, we have implemented an INT8 KV cache approach. To ensure that the INT8 representation closely approximates higher precision data types, we maintain a unique scale for each token and each head as shown in Figure ~\ref{kvcache1}. 
In this method, each head has a unique scale value, which could regulate the KV cache in a fine grained pattern, while with acceptable storage increasing.
This approach allows for a more efficient use of memory without significantly compromising the quality of the model's output. Moreover, we engineered a custom kernel capable of supporting MatMul operations with hybrid data types. This kernel is adept at handling INT8 data, which it dynamically converts to FP32 during execution, as shown in Figure ~\ref{kvcache2}. This conversion process is integral to leveraging the Fused Multiply-Add(FMA) instructions facilitated by the AVX512 instruction set, a feature of the most recent x86 architectures. The conversion from INT8 to FP32 is a two-step process: initially, the \_mm512\_cvtepi8\_epi32 intrinsic function \cite{intel2024intrinsics} transforms INT8 into INT32, followed by the \_mm512\_cvtepi32\_ps function, which then converts the INT32 data into FP32 format. This approach ensures efficient utilization of the available instruction set for optimized computational throughput.

\subsection{Distributed Inference Optimization}

To enhance distributed inference performance, we present a solution for optimizing distributed inference for LLMs on CPUs. This solution is implemented using the oneAPI Collective Communications Library (oneCCL).

\begin{figure}[h]
	\begin{center}
		\includegraphics[width=0.5\textwidth]{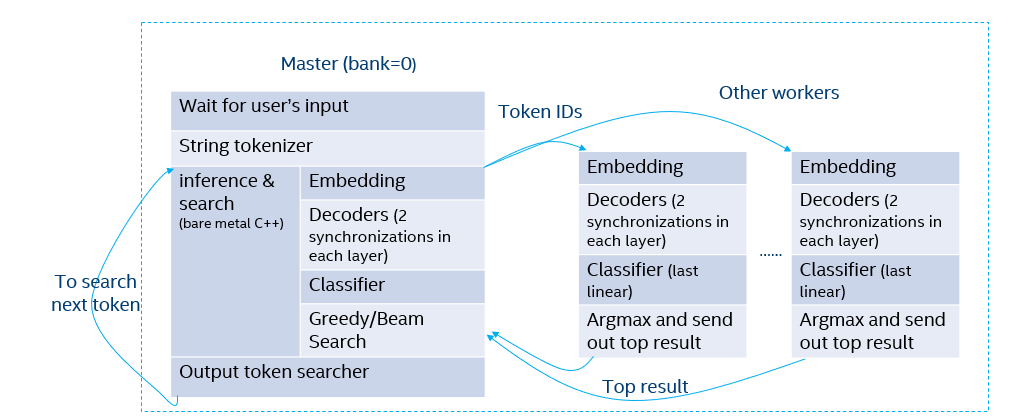}
	\end{center}
	\caption{Distributed inference based on oneCCL.}
	\label{dis1}
\end{figure}

In the proposed solution, our solution broadcasts token IDs
instead of broadcasting the values of the Embedding part obtained based on token IDs.
We perform the reduction after each worker computes the top-k, rather than directly reducing the logits of all tokens.
The implementation is shown in Figure ~\ref{dis1} .

\begin{figure}[h]
	\begin{center}
		\includegraphics[width=0.4\textwidth]{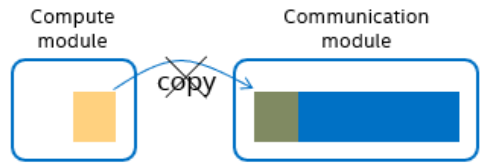}
	\end{center}
	\caption{A zero-copy implementation}
	\label{dis3}
\end{figure}

Moreover, an aggressive optimization approach was proposed for 
reduce data copying since we found that when the computation module and communication module interact, data copying is often involved in practice.
This involves the computation module, during its last operation before communication, directly writing the results to the location of the communication module, achieving a zero-copy implementation which shown in Figure ~\ref{dis3}.

\section{Experiment Results}
\label{exp}

\subsection{Configuration}
	
	We conducted experiments on the Intel$^\circledR$ Xeon$^\circledR$ CPU 8563C, and detailed information of one bare-mental machine is provided in Table ~\ref{hardware}. Note that each machine has 2 sockets. 
	\begin{table}[t]
		\caption{Hardware configuration}
		\label{hardware}
		\begin{center}
			\begin{tabular}{lc}
				\multicolumn{1}{c}{\bf HEADINGS}  &\multicolumn{1}{c}{\bf CONFIG}
				\\ \hline \\
				CPU Model        & Intel Xeon CPU 8563C \\
				Sockets          & 2                       \\
				Cores per Socket & 52                      \\
				Base Frequency   & 2.6GHz                  \\
				All-core Max Frequency & 3.1GHz  
			\end{tabular}
		\end{center}
	\end{table}

\subsection{Performance}
	
We measure the throughput and latency of next token generation on Intel$^\circledR$ Xeon$^\circledR$ CPU 8563C by adopting the proposed solution. 

Table \ref{perf dis} shows next token generation latency results of the Llama2 \cite{touvron2023llama} with 70B parameters under a model configuration with input tokens = 1024, output tokens = 128, batch size = 1. The latency result is better when it is lower.
With 4 bare-mental machines (8 sockets) running the proposed distributed solution, it could bring 2.85X performance gain for Llama2-70B compared with 1 bare-mental machine (2 sockets).

	\begin{table}[t]
		\caption{Next token generation latency of Llama2-70B with the proposed distributed solution}
		\begin{center}
			\begin{tabular}{ll}
				\multicolumn{1}{c}{\bf SOCKET}  &\multicolumn{1}{c}{\bf LATENCY}
				\\ \hline \\
				2    &   249.7 ms \\
				8    &   87.7 ms
			\end{tabular}
			\label{perf dis}
		\end{center}
	\end{table}
	
Table \ref{llm opt} shows the performance comparison of SlimAttention and FlashAttention on 1 socket. 
In table \ref{llm opt}, 1st column is input length, 2nd column is the performance of FlashAttention while 3rd column is the performance of the proposed SlimAttention.
Note that performance refers to the average cost time (ms) of each attention layer during the generation of the first token with configuration of Llama2-7B model when batch size = 1. It shows the proposed SlimAttention approach could achieve better performance on CPU.

\begin{table}[h!]
	\centering
	\begin{tabular}{lll}
		\textbf{INPUT} & \textbf{FLASH} & \textbf{PROPOSED SLIM} \\
		\hline
		256 & 10.85 & 1.10 \\
		512 & 27.95 & 6.60 \\
		1024 & 61.57 & 16.02 \\
		2048 & 176.36 & 96.65 \\
		4096 & 540.14 & 392.80 \\
	\end{tabular}
	\caption{Performance comparison between FlashAttention and SlimAttention for different input token sizes.}
	\label{llm opt}
\end{table}

Table \ref{perf through} shows throughput without first token of Llama2-7B on 1 socket, under a model configuration with input tokens = 148, output tokens = 198, and the batch size is 256/512.  
Throughput result is better when it is higher. The proposed solution demonstrates its potential to enhance throughput effectively. 

\begin{table}[t]
	\caption{Throughput without first token of Llama2-7B}
	\begin{center}
		\begin{tabular}{ll}
			\multicolumn{1}{c}{\bf BATCH SIZE}  &\multicolumn{1}{c}{\bf LATENCY}
			\\ \hline \\
			256    &   796.9 tokens/s \\
			512    &   853.6 tokens/s
		\end{tabular}
		\label{perf through}
	\end{center}
\end{table}

\section{Conclusion and Future Work}
	
We presented an end-to-end LLM inference accelerating solution including distributed inference optimization, effective KV cache optimization, and individual LLM optimization. 
We showcased the versatility across a range of popular LLMs and the performance superiority over the open-source solution on CPUs.
In our forthcoming research, we intend to broaden our study to include a wider variety of CPUs, particularly those with resource constraints. 
Our primary focus will be on enhancing performance for larger batch sizes and exploring effective deployment serving solutions. 
Concurrently, we aim to adapt our solution to accommodate the latest trending models, such as the mixture of experts (MoE) models. Our goal is to offer a practical alternative to existing GPU solutions.

\bibliography{xft}

\begin{thebibliography}{11}
\providecommand{\natexlab}[1]{#1}
\providecommand{\url}[1]{\texttt{#1}}
\expandafter\ifx\csname urlstyle\endcsname\relax
  \providecommand{\doi}[1]{doi: #1}\else
  \providecommand{\doi}{doi: \begingroup \urlstyle{rm}\Url}\fi

\bibitem[Bai et~al.(2023)Bai, Bai, Chu, Cui, Dang, Deng, Fan, Ge, Han, Huang,
  et~al.]{bai2023qwen}
Bai, J., Bai, S., Chu, Y., Cui, Z., Dang, K., Deng, X., Fan, Y., Ge, W., Han,
  Y., Huang, F., et~al.
\newblock Qwen technical report.
\newblock \emph{arXiv preprint arXiv:2309.16609}, 2023.

\bibitem[Dao et~al.(2022)Dao, Fu, Ermon, Rudra, and
  R{\'e}]{dao2022flashattention}
Dao, T., Fu, D., Ermon, S., Rudra, A., and R{\'e}, C.
\newblock Flashattention: Fast and memory-efficient exact attention with
  io-awareness.
\newblock \emph{Advances in Neural Information Processing Systems},
  35:\penalty0 16344--16359, 2022.

\bibitem[Intel(2024)]{intel2024intrinsics}
Intel, R.
\newblock Intrinsics guide, 2024.
\newblock
  \url{https://www.intel.com/content/www/us/en/docs/intrinsics-guide/index.html}.

\bibitem[Liu et~al.(2023)Liu, Lin, Hewitt, Paranjape, Bevilacqua, Petroni, and
  Liang]{liu2023lost}
Liu, N.~F., Lin, K., Hewitt, J., Paranjape, A., Bevilacqua, M., Petroni, F.,
  and Liang, P.
\newblock Lost in the middle: How language models use long contexts.
\newblock \emph{arXiv preprint arXiv:2307.03172}, 2023.

\bibitem[Thirunavukarasu et~al.(2023)Thirunavukarasu, Ting, Elangovan,
  Gutierrez, Tan, and Ting]{thirunavukarasu2023large}
Thirunavukarasu, A.~J., Ting, D. S.~J., Elangovan, K., Gutierrez, L., Tan,
  T.~F., and Ting, D. S.~W.
\newblock Large language models in medicine.
\newblock \emph{Nature medicine}, 29\penalty0 (8):\penalty0 1930--1940, 2023.

\bibitem[Touvron et~al.(2023)Touvron, Martin, Stone, Albert, Almahairi, Babaei,
  Bashlykov, Batra, Bhargava, Bhosale, et~al.]{touvron2023llama}
Touvron, H., Martin, L., Stone, K., Albert, P., Almahairi, A., Babaei, Y.,
  Bashlykov, N., Batra, S., Bhargava, P., Bhosale, S., et~al.
\newblock Llama 2: Open foundation and fine-tuned chat models.
\newblock \emph{arXiv preprint arXiv:2307.09288}, 2023.

\bibitem[Vaswani et~al.(2017)Vaswani, Shazeer, Parmar, Uszkoreit, Jones, Gomez,
  Kaiser, and Polosukhin]{vaswani2017attention}
Vaswani, A., Shazeer, N., Parmar, N., Uszkoreit, J., Jones, L., Gomez, A.~N.,
  Kaiser, {\L}., and Polosukhin, I.
\newblock Attention is all you need.
\newblock \emph{Advances in neural information processing systems}, 30, 2017.

\bibitem[Wu et~al.(2023)Wu, He, Liu, Sun, Liu, Han, and Tang]{wu2023brief}
Wu, T., He, S., Liu, J., Sun, S., Liu, K., Han, Q.-L., and Tang, Y.
\newblock A brief overview of chatgpt: The history, status quo and potential
  future development.
\newblock \emph{IEEE/CAA Journal of Automatica Sinica}, 10\penalty0
  (5):\penalty0 1122--1136, 2023.

\bibitem[Yang et~al.(2023)Yang, Xiao, Wang, Zhang, Bian, Yin, Lv, Pan, Wang,
  Yan, et~al.]{yang2023baichuan}
Yang, A., Xiao, B., Wang, B., Zhang, B., Bian, C., Yin, C., Lv, C., Pan, D.,
  Wang, D., Yan, D., et~al.
\newblock Baichuan 2: Open large-scale language models.
\newblock \emph{arXiv preprint arXiv:2309.10305}, 2023.

\bibitem[Zhang et~al.(2022)Zhang, Roller, Goyal, Artetxe, Chen, Chen, Dewan,
  Diab, Li, Lin, et~al.]{zhang2022opt}
Zhang, S., Roller, S., Goyal, N., Artetxe, M., Chen, M., Chen, S., Dewan, C.,
  Diab, M., Li, X., Lin, X.~V., et~al.
\newblock Opt: Open pre-trained transformer language models.
\newblock \emph{arXiv preprint arXiv:2205.01068}, 2022.

\bibitem[Zhao et~al.(2023)Zhao, Zhou, Li, Tang, Wang, Hou, Min, Zhang, Zhang,
  Dong, et~al.]{zhao2023survey}
Zhao, W.~X., Zhou, K., Li, J., Tang, T., Wang, X., Hou, Y., Min, Y., Zhang, B.,
  Zhang, J., Dong, Z., et~al.
\newblock A survey of large language models.
\newblock \emph{arXiv preprint arXiv:2303.18223}, 2023.

\end{thebibliography}
\bibliographystyle{icml2024}



\end{document}